\documentclass{article} % For LaTeX2e
\usepackage{nips13submit_e,times}
\usepackage{hyperref}
\usepackage{url}
\usepackage{graphicx}
\usepackage{multirow}
\usepackage[numbers,sort&compress]{natbib}
\usepackage{mathtools}
\usepackage{amsmath}
\usepackage{algorithm2e}

\title{Adversarial Image Alignment and Interpolation}

\author{
Viren Jain \\
Google Research\\
\texttt{viren@google.com} \\
}

\nipsfinalcopy 

\begin{document}

\maketitle

\begin{abstract}
Volumetric (3d) images are acquired for many scientific and biomedical purposes using imaging methods such as serial section microscopy, CT scans, and MRI. A frequent step in the analysis and reconstruction of such data is the alignment and registration of images that were acquired in succession along a spatial or temporal dimension. For example, in serial section electron microscopy, individual 2d sections are imaged via electron microscopy and then must be aligned to one another in order to produce a coherent 3d volume. State of the art approaches find image correspondences derived from patch matching and invariant feature detectors, and then solve optimization problems that rigidly or elastically deform series of images into an aligned volume. Here we show how fully convolutional neural networks trained with an adversarial loss function can be used for two tasks: (1) synthesis of missing or damaged image data from adjacent sections, and (2) fine-scale alignment of block-face electron microscopy data. Finally, we show how these two capabilities can be combined in order to produce artificial isotropic volumes from anisotropic image volumes using a super-resolution adversarial alignment and interpolation approach. 
\end{abstract}

\section{Introduction}

State of the art methods for many common image processing tasks such as classification, detection, and segmentation are based on supervised machine learning, and in recent years multi-layer neural networks have been one of the main drivers of progress. However, there remain certain tasks which have been difficult to solve using machine learning tools, often due to the lack of any appropriate target label dataset or other supervisory signal. For example, in the task of image registration and alignment, it is generally difficult to collect large scale human ground truth data due to the cumbersome nature of manually performing the precise image manipulations involved in an affine or elastic deformation of a series of images. Therefore the parameters involved in image alignment pipelines are typically hand-tuned to deliver satisfactory results, rather than automatically optimized based on ground truth labels.\footnote{While the results from these hand-tuned pipelines could in principle serve as a labeled dataset for training machine learning methods, this approach would be very unlikely to lead to learned methods that perform better or differently.}

In this work we explore a novel approach to the problem of image interpolation and image alignment based on two core tools: feedforward neural networks and adversarial loss functions. The primary contributions of this work are:

\begin{itemize}
\item A convolutional network architecture that can interpolate an image section from adjacent section or produce transformed versions of volume data that are better aligned. 
\item An unsupervised, adversarial learning criteria for the problem of image alignment and image interpolation.
\item A neural network architecture that combines interpolation and alignment in order to produce `super-resolution' versions of  anisotropic data. 
\item An experimental evaluation of these approaches on volumetric, block-face scanning electron microscopy images of brain tissue. 
\end{itemize}

\section{Related Work}
Registering, stitching, and aligning images that represent contiguous samples along some axis in space or time is a computer vision problem that has found  applications across photography, microscopy, astronomy, and many other areas \cite{szeliski2010computer}. Classical approaches make heavy use of image-derived correspondences computed by local correlation measurements or robust interest point detectors \cite{Lowe:2004}, which are used to estimate how overlapping or nearly overlapping images geometrically relate to one another. These estimates can then be used to, for example, define an optimization problem in which images are elastically deformed to maximize pixel-wise consistency \cite{saalfeld2012elastic}. Such approaches have proven highly effective for producing coherent 3d volumes from thousands or millions of individual 2d images that may be subject to various types of deformation, artifacts, and missing data. 

Heinrich and colleagues recently introduced neural network architectures that produce impressive results on the task of generating isotropic 3d volumes from anisotropic source data \cite{heinrich2017deep}, which is closely related to the experiments we pursue in Section 5. The novel training methods introduce in this work are likely to be complementary to general advances in neural network architectures for 3d image synthesis.

Finally, automated synthesis of image data has recently been advanced by the invention of generative adversarial networks (GANs) \cite{goodfellow2014generative, radford2015unsupervised}. These approaches enable a new approach to expressing loss-functions for many different unsupervised modeling tasks in which previous approaches produced results that were generally unconvincing to human observers. Hundred of papers in recent years have applied the GAN approach to tasks such as producing class-specific image data and synthesizing frames in video, as well as some preliminary attempts at more discrete tasks such as natural language generation. In this work, we show how to apply some of these methods to the problems of image interpolation, alignment, and super-resolution. 

\section{Adversarial Section Interpolation}

\begin{figure}
    \centering
    \includegraphics[scale=0.4]{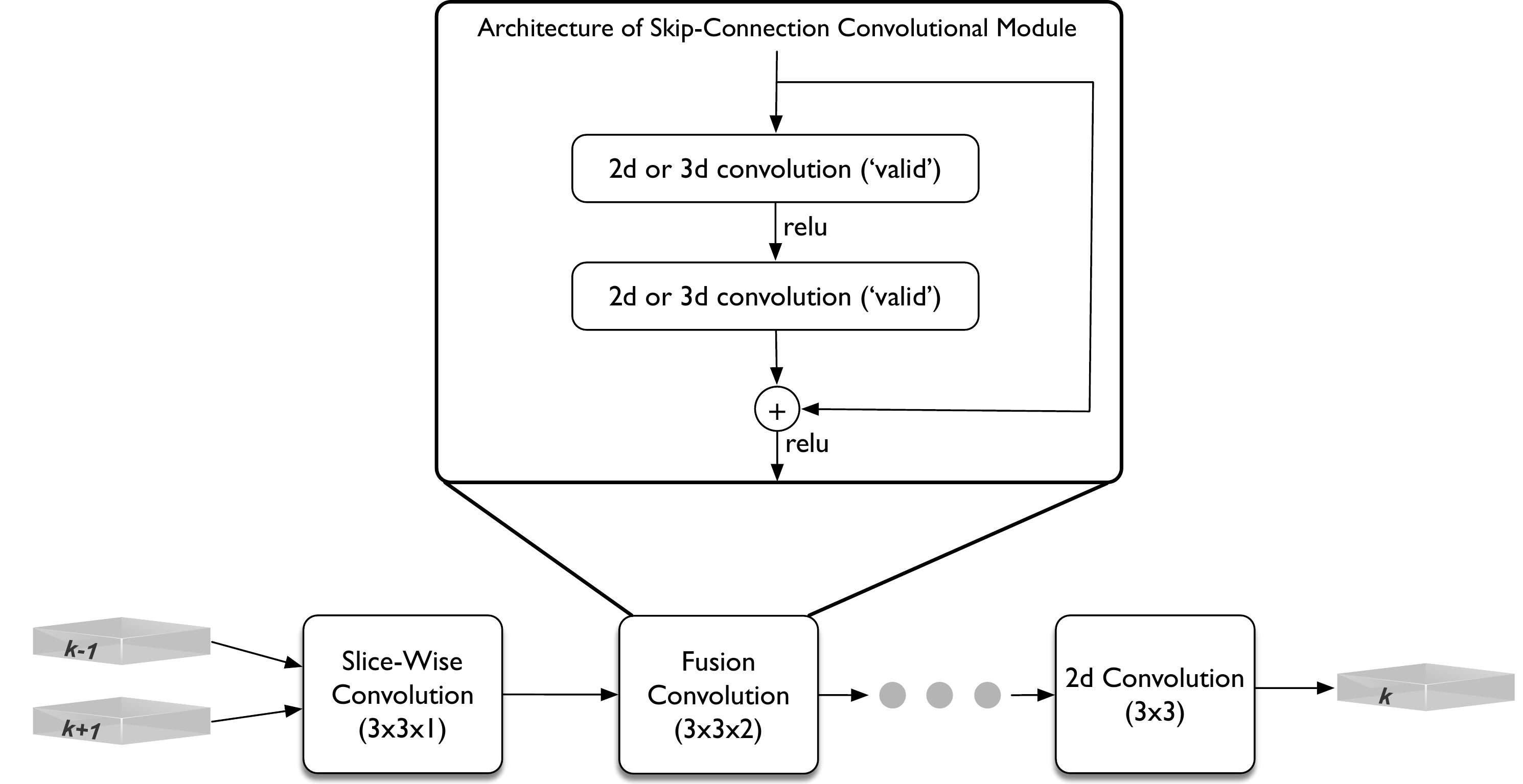}
    \caption{Architecture of the slice-interpolation network.}
    \label{fig:architecture_slice_interpolation}
\end{figure}

We begin with the problem of interpolation of individual 2d image sections within 3d image volumes. In particular, we address the problem of synthesizing image data for a particular section $k$ within some (already aligned) 3d image volume, given data from neighboring sections $k-1$ and $k+1$. This problem has practical relevance in, for example, the analysis of 3d microscopy stacks where there may be an artifact in one slice that deteriorates the ability to manually or automatically analyze the volume as a continuous set of serial sections. In some cases, a section may be entirely missing due to acquisition issues, and such an interpolation algorithm could be used to `hallucinate' the missing slice. 
We note that the baseline (non-adversarial) approach to solving section interpolation with neural networks offers a completely supervised learning problem (since an image stack most likely contains many examples of completely valid $\{k-1, k, k+1\}$  section tuples). However, we demonstrate that augmenting the supervised learning problem with an (unsupervised) adversarial loss function can significantly improve results. 

\subsection{Network Architecture}
We intentionally focus on a relatively simple architecture in order to demonstrate that simple networks are capable of accurate interpolation, if trained with an adversarial loss function. 

The input to the network is two image sections: the neighboring $k-1$ and $k+1$ slice for the desired output slice $k$. The $x$ and $y$ extent of the images is somewhat arbitrary due to the fully convolutional structure of the network; during training, we used $100 \times 100$. The network consists of three phases: two convolution modules with slice-wise 2d convolutions (filter size $3 \times 3 \times 1$ in `valid' mode), followed by a single 3d convolution that fuses the two `feature map sections' into a single section (filter size $3 \times 3 \times 2$ in 'valid mode'), followed by three convolution modules with 2d convolutions that output the final interpolated section (filter size $3 \times 3$). Within each convolution module, we use skip connections (similar to the residual units of \cite{he2015deep}); specifically, we used an architecture equivalent to ``full pre-activation'' of \citet{he2016identity}. The final convolution in the network uses a linear activation function in order to permit reproduction of the zero mean and unit standard deviation target values. All layers, except for the output, have fifty filters (feature maps). No pooling is used anywhere in the network (see Figure~\ref{fig:architecture_slice_interpolation}). 

\subsection{Baseline: Pixel-Wise Training}

In the baseline (non-adversarial) case, the network is trained using only explicit pixel-wise target values as given by some `ground truth' $k$th slice. Both the input and target images were normalized to zero mean and unit standard deviation. We experimented with both a mean squared error loss function as well as absolute difference, and found that using an absolute difference error signal produced slightly better results (based on visual evaluation). We implemented the model in TensorFlow \cite{abadi2016tensorflow} and trained it with asychronous gradient descent using eight NVIDIA K20 GPU. The learning rate was set to $0.001$, and the batch size to 6.

\subsection{Adversarial Training}
\begin{algorithm}
 \While{$step \leq max\_step$}{
  Sample minibatch $\{ x_1, x_2, .., x_m \}$ where $x_i=\{ SLICE_{k-1}, SLICE_{k}, SLICE_{k+1} \}$\;
  Sample minibatch $\{ y_1, y_2, .., y_m \}$ where $y_i=\{ SLICE_{k-1}, SLICE_{k} \}$\;
  Run generator $G$ to produce interpolated slices: $G_{x_i} = G(\theta_G; x_i)$ \;
  Run discriminator $D$ on `fake pair:' $D_{x_i} = D(\theta_D; \{SLICE_{k-1} \in x_i,  G_{x_i}\})$\;
  Run discriminator on `real pair:' $D_{y_i} = D(\theta_D; y_i)$\;
  $D_{LOSS} = \sum_{i=1}^m L(D_{x_i}, 0) + \sum_{i=1}^m L(D_{y_i}, 1)$\;
  $G_{LOSS} = \sum_{i=1}^m L(D_{x_i}, 1)$\;
  \If{ $use\_pixelwise\_loss$ }{
   $G_{LOSS} = G_{LOSS} + \sum_{i=1}^m L_{PIXEL}(G_{x_i}, x_i)$\;
   }
  Update discriminator: $\theta_D \coloneqq \theta_D - \eta \nabla_{\theta_D}D_{LOSS}$\;
  Update generator: $\theta_G \coloneqq \theta_G - \eta \nabla_{\theta_G}G_{LOSS}$\;
  Update generator: $\theta_G \coloneqq \theta_G - \eta \nabla_{\theta_G}G_{LOSS}$\;
  $step=step+1$;
 }
 \caption{Training loop for the adversarial slice-interpolation network. $\theta_D$ and $\theta_G$ are trainable parameters associated with the discriminator and generator networks respectively. The generator update is performed twice. Note that in practice we use the ADAM optimizer \cite{adam} but here we present a simplified description with standard stochastic gradient descent. $L_{PIXEL}$ is the sum over pixel-wise absolute differences, which in our experiments we found performed better than mean squared error. }
 \label{algorithm:adversarial_slice_interpolation}
\end{algorithm}

We implemented an adversarial approach to training the slice-interpolation networks \cite{goodfellow2014generative, radford2015unsupervised}. We refer to the slice-interpolation network as the \emph{generator} and a separate convolution-pooling network as the \emph{discriminator}. The $k-1$ and $k+1$ slice from some image volume provides input to the generator, which outputs an interpolated slice $k$. The discriminators task is to distinguish whether a pair of slices (consisting of slice $k-1$ and \emph{either} slice $k$ or interpolated slice $k$) is `fake' (including the slice $k$ given by the generator) versus `real' (including the slice $k$ sampled from the true data). 

The architecture of the discriminator is a straightforward 2d convolution-pooling network: three modules consisting of alternating convolution (filter size $5 \times 5$) and max-pooling (kernel and stride of $2 \times 2$) that are applied to each of the two slices independently, followed by a `flattening' layer that concatenates the information from both slice $k-1$ and $k$, followed by dropout, a fully-connected layer, and finally a sigmoid scalar output. By back-propagating the fake versus real classification error signal from the discriminator through the generator, the generator learns to produce output that can `fool' the discriminator into believing the interpolated section is real. A subtlety in this training scheme lies in the balance between the discriminator and generator; if the discriminator is too weak, it cannot force the generator to change its behavior. If the generator is too weak, then it may not be able to usefully exploit the discriminators error signal to achieve a better solution. In Algorithm~\ref{algorithm:adversarial_slice_interpolation} we provide pseudocode for the adversarial training scheme. In our implementation we used the ADAM optimizer \cite{adam} with a learning rate of $0.002$ and $\beta_1$ set to $0.5$. The batch size was 6 ($m=6$ in the notation defined in Algorithm~\ref{algorithm:adversarial_slice_interpolation}). We implemented the model in TensorFlow and trained on a single NVIDIA K20 GPU. 

\subsection{Serial Block Face Section Interpolation Experiments}
\begin{figure}
    \centering
    \includegraphics[scale=0.7]{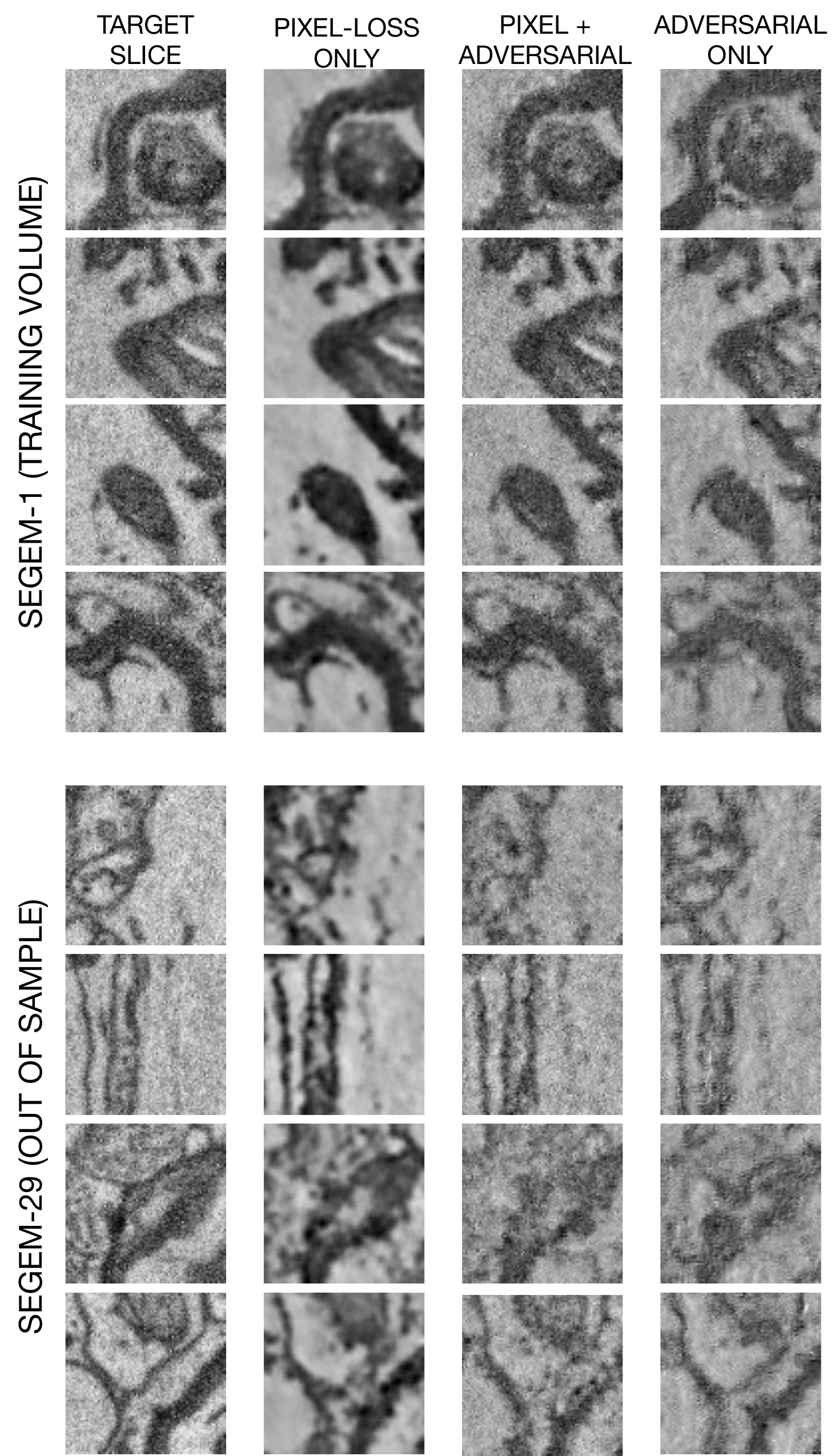}
    \caption{Adversarial slice interpolation experiments on serial block face volumes of mouse cortex (SegEM).}
    \label{fig:experiments_slice_interpolation_segem}
\end{figure}
We performed section interpolation experiments on images of mouse S1 neocortex imaged at 11.24 $\times$ 11.24 $\times$ 28 $nm^3$ with Serial Block-face Electron Microscopy (SBEM), released as part of the SegEM project \cite{berning2015segem}. Specifically, we used `volume 1' to train the models and `volume 29' as an out-of-sample test (each volume is 200 $\times$ 200 $\times$ 150 voxels). 

Figure~\ref{fig:experiments_slice_interpolation_segem} compares the results of the adversarial approach to the baseline pixel-loss only strategy. The adversarial loss dramatically improves the plausibility of the interpolated sections. Interestingly, the adversarial network produces a reasonably convincing `simulation' of the shot noise that characterizes samples of the real data, as well as much sharper and more realistic ultrastructural boundaries as compared to the relatively smoothed out images generated by the network trained only with pixel-loss.  

The model produces better results on the `training' volume as compared to the  out-of-sample volume. However, we note that in this case the training is unsupervised, so in practice the training strategy can be deployed on the dataset where one is directly interested in applying the model. Furthermore, the training volume in this experiment is very small relative to real connectomic datasets; in other experiments with $1000\times$ larger training data (i.e., full connectomic volumes), we have observed no significant generalization issues. 

\section{Adversarial Alignment}
The goal of volumetric image alignment is to produce a single 3d volume from multiple 2d images where physically coherent structures are pixel-wise contiguous in all directions (to the extent permitted by the isotropy or anisotropy of the underlying imagery). In general, it has been difficult to define a completely reliable quantitative objective function for this task. For example, a naive approach would be to aggressively maximize pixel-wise correlation from one image to the next. In this case, however, a perfect solution would be given by the trivial and useless solution of making each image exactly identical. Therefore, highly regularized optimizations have been `hand-crafted' to maximize image-to-image consistency without losing important details and variability in the images \cite{saalfeld2012elastic}. 

Human evaluation of image alignment quality is often performed by visual inspection of an 2d orthogonal `reslice' of the volume data ($xz$ or $yz$ planes). The underlying assumption is that after successful alignment the boundaries and edges of physically contiguous structures should appear smooth in the reslice; or, more generally, the \emph{reslice should appear similar to the imaging plane}, assuming no preferred orientation of the underlying image content and roughly isotropic pixel resolution.
\begin{figure}
    \centering
    \includegraphics[scale=0.7]{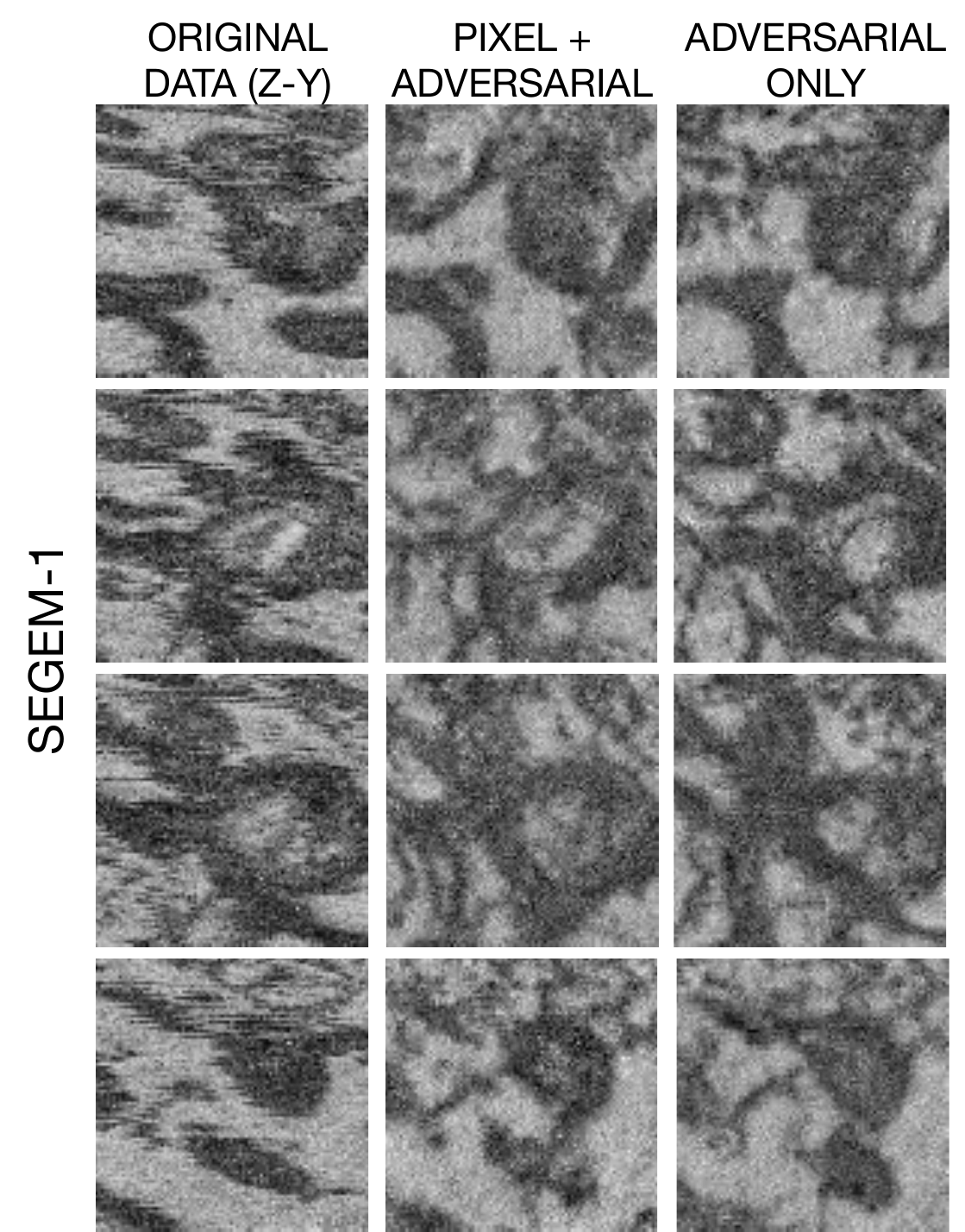}
    \caption{Adversarial alignment experiments on serial block face volumes of mouse cortex (SegEM). Each image is a $91\times91$ $zy$ reslice from either the original data, output of a network trained with both adversarial and pixel loss, or output of a network trained only with adversarial loss.}
    \label{fig:experiments_adversarial_alignment_segem}
\end{figure}
We therefore tested an adversarial approach to image alignment in which the discriminators goal was to distinguish $xy$ (imaging plane) slices versus post-aligned $yz$ and $xz$ reslices. The generator is trained to fool the discriminator, which we hypothesized would result in the generator network learning to align the data. 

\subsection{Network Architecture and Training}
As in our experiments on section interpolation, we intentionally focused on an extremely simple network architecture for both the generator and discriminator. The generator was a 3d fully convolutional network with $3\times3\times3$ kernels and skip connections (nearly identical to the architecture in Figure~\ref{fig:architecture_slice_interpolation} except with only 3d `valid' convolutions). The input to the network is a full 3d stack of unaligned data, and the output is the `post-aligned' 3d volume of similar dimensions (modulo valid convolution effects). The discriminator is the same 2d conv-pooling network described in Section 3.3 that receives as input either an $xy$ slice form the (real) input data, or an $xz$ or $yz$ reslice from the output of the generator. The $xy$ slices are cropped to match the size of the orthogonal reslices so that the discriminator network receives identically sized images in all cases. 

The training algorithm is identical to that described in ~\ref{algorithm:adversarial_slice_interpolation} with slight alterations for the fact that the input to the discriminator is in this case just either a single `real' slice from the input minibatch or a `fake' post-aligned re-slice. 

% https://gfsviewer.corp.google.com/cns/ok-d/home/viren/adversarial_alignment_pngs/aligner/segem1_z_y_slice_60_100_100_50_actual.png

\subsection{Serial Block Face Adversarial Alignment Experiments}
We performed adversarial alignment experiments with the S1 neocortex data described in Section 3.4. The data is roughly aligned, however, there remains significant slice-to-slice jitter at the pixel-level that is visible in $xz$ or $yz$ slices of the original data.

A subtlety to the approach lies in whether to use a pixel-wise cost function (e.g., mean squared error or absolute difference) in addition to the adversarial loss. In this case (and unlike section interpolation), the pixel-wise loss is clearly wrong, in that we seek a new post-aligned volume with different pixel values from the original volume. However, we also want the network to preserve important details in the original image, not just generate `plausible' images that might, for example, be missing a synaptic vesicle or other crucial image content. Therefore we tested the adversarial alignment both with and without a pixel-wiss loss.

Figure~\ref{fig:experiments_adversarial_alignment_segem} compares a reslice of the original volume (`data z-y') to a reslice of the data after applying a network trained with pixel and adversarial loss functions. Visual inspections suggests that that the post-processed data indeed appear more similar to the imaging plane (examples shown in Figure~\ref{fig:experiments_slice_interpolation_segem}) and the jitter and discontinuities obvious in the input data are noticeably reduced. 

\section{Super-Resolution Adversarial Alignment and Interpolation}
The alignment experiments in section 4.2 used data imaged at 11.24 $\times$ 11.24 $\times$ 28 $nm^3$, or roughly double the voxel size in $z$ compared to $xy$. Hence there are clear limits to the extent to which a post-aligned orthogonal reslice can ever be visually similar to the imaging plane (without introducing potentially dramatic distortions to the image content). Therefore we combined the adversarial section interpolation and alignment methods into a unified approach which outputs upsampled volumetric data that seeks to both `correct' for the anisotropy of the original data and align the data.

\subsection{Network Architecture and Training}
\begin{figure}
    \centering
    \includegraphics[scale=0.4]{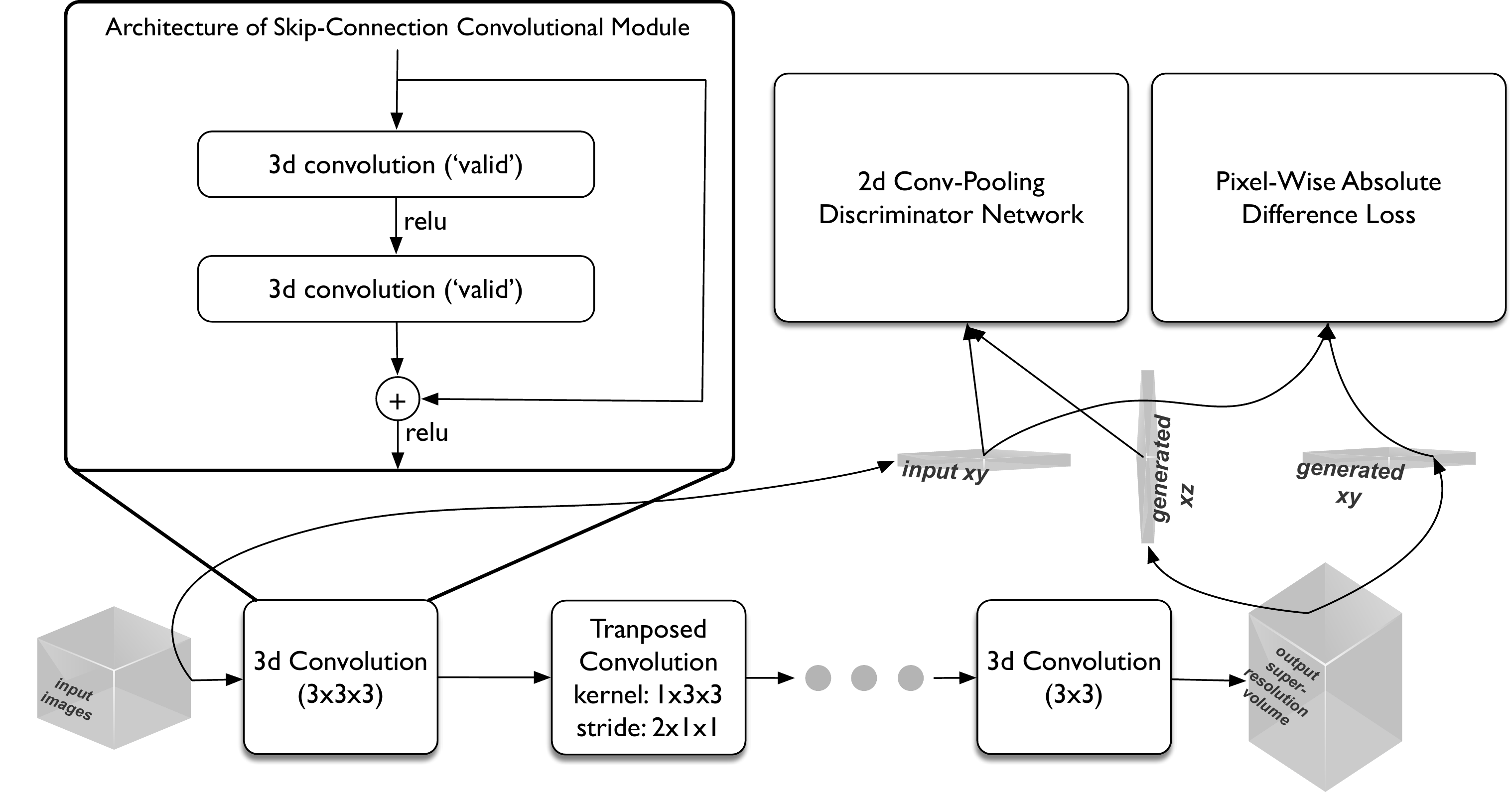}
    \caption{Architecture of the super-resolution adversarial alignment and interpolation network.}
    \label{fig:architecture_super_resolution}
\end{figure}

The network architecture of the generator follows the one described in Section 4.1, except with an intermediate stage that uses `transposed' 3d convolution to upsample the data by $2\times$ in the $z$ axis \cite{dumoulin2016guide}. Specifically, the architecture has three modules of skip-connection 3d convolution ($3\times3\times3$ kernels), followed 3d transposed convolution (kernel size of $1\times3\times3$ with stride $2\times1\times1$) followed by two additional modules of skip-connection 3d convolution ($3\times3\times3$ kernels). All convolutions are performed in `valid' mode. The network architecture outputs volumes that are (in pixel dimensions) exactly $2\times$ as large in $z$ as compared to $x$ (for isotropic input volume dimensions). 

The discriminator follows the 2d convolution-pool structure established in the prior experiments. However, due to the difference in pixel dimensions of the $xz$ or $yz$ reslice versus the $xy$ imaging plane arising from the upsampling, we crop the reslice planes to a size equivalent to the $xy$ size so that the discriminator receives the same image size in all cases. The training algorithm is identical to that used for adversarial alignment. 

\subsection{Experiments}
\begin{figure}
    \centering
    \includegraphics[scale=0.7]{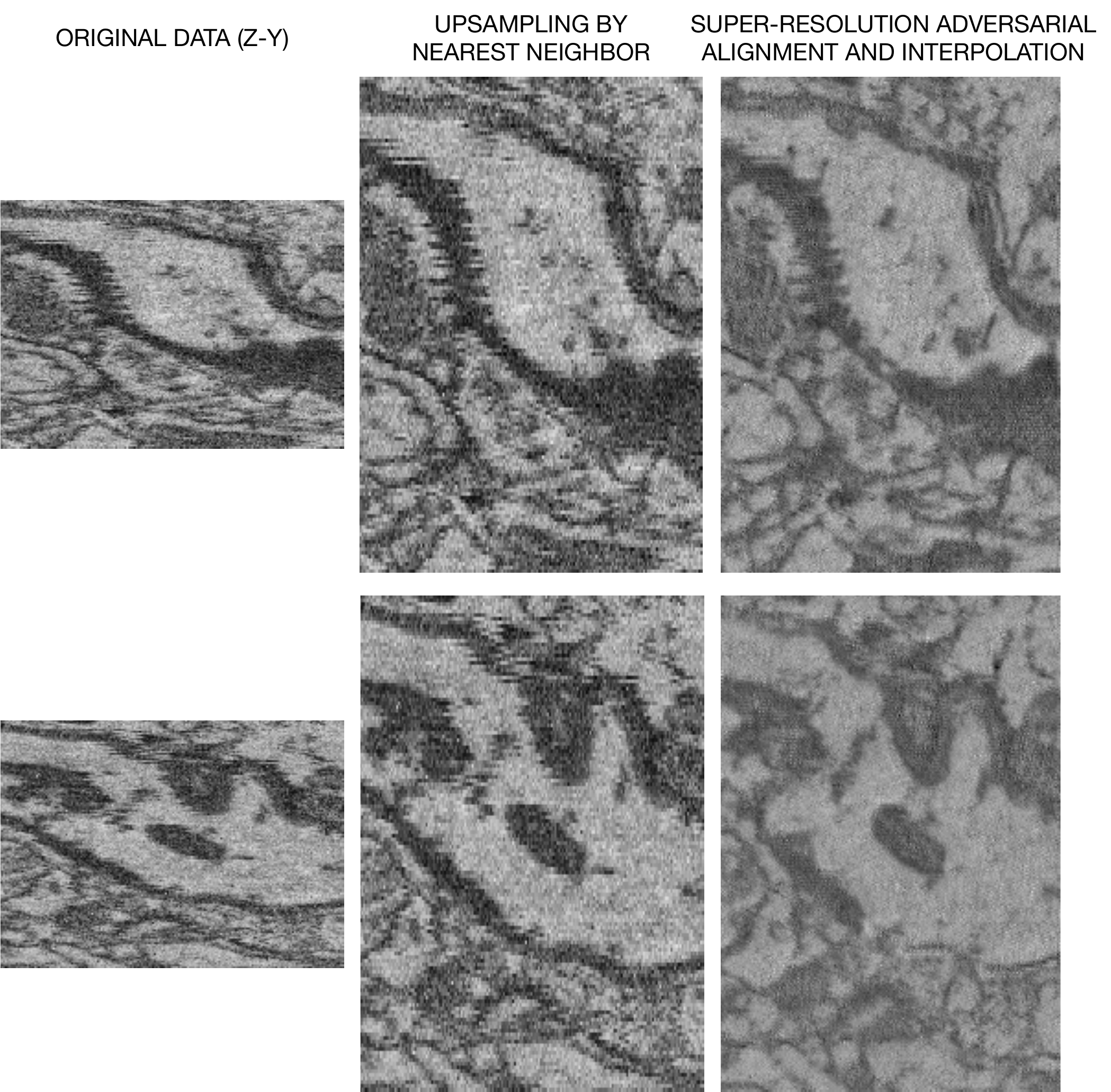}
    \caption{Super-resolution adversarial alignment and interpolation experiments on serial block face volumes of mouse cortex. Left side (original data) images are $135\times185$ pixels, and right side (network output) are $270\times185$ pixels.}
    \label{fig:experiments_super_resolution_segem}
\end{figure}

We performed super-resolution adversarial alignment and interpolation on the S1 neocortex data described in section 3.4. We applied an absolute difference pixel loss to alternating slices in the generative model output (since the original data can only be matched to half the slices in the super-resolution output of the network). Figure~\ref{fig:experiments_super_resolution_segem} compares the original data reslice to the super-resolution adversarial alignment and interpolation output.  

\section{Discussion}

We have presented a novel approach to the problem of interpolating individual images within a series, achieving a pixel-wise `fine-scale' alignment of volumetric data, and performing super-resolution alignment and interpolation of volumetric data. Overall, we regard the described approach and results as an initial step in a new direction. Further work is likely required before these methods can be considered a true alternative to established best practice; we conclude by describing several such directions.

In this paper we focused on highly simple network architectures to isolate the effects of the adversarial learning approach from improvements in image synthesis and modeling that can be achieved purely by utilizing more advanced neural networks. U-Nets and other new architectures that have recently been exploited for tasks such as super-resolution will likely improve the power of the methods proposed here \cite{heinrich2017deep}.

Adversarial learning itself is a rapidly evolving field where there is little consensus regarding best practice for maximizing the stability and efficacy of learning procedures. Indeed, recent advances suggest various improvements over the methods used in the experiments described here \cite{arjovsky2017wasserstein}. Moreover, if these approaches are to be deployed at a large scale in the context of a challenging end-to-end machine perception task such as connectomic reconstruction, it will be important to understand the extent to which such methods can capture the full data distribution and when they might fail \cite{arora2017generalization}.

Finally, for a task such as image alignment, it is reasonable to question the entire approach of using a pixel-level neural network to simply generate new pixel values from an input serial section series. For example, it may be difficult to formulate such an approach that could efficiently model displacements of tens to hundreds of pixels, which are generally required in the initial stages of alignment of highly anisotropic datasets. Furthermore, the neural network provides no guarantees as to which elements and features of the input data are preserved in the `aligned' output, which complicates analysis of pipelines that perform downstream annotation or segmentation. Therefore, one may ultimately desire a `hybrid' approach in which the various parameters of the more constrained transforms used in current large-scale alignment pipelines are automatically optimized using an adversarial objective function. 

\bibliographystyle{plainnat}
\bibliography{alignment}

\end{document}